\documentclass[12pt]{article}
\usepackage{amsfonts}
\usepackage{amsmath}
\usepackage{url}
\usepackage{stmaryrd}
\usepackage[all,cmtip]{xy}

\newcommand {\CC}{\mbox{${\mathcal{C} }$}}

\begin{document}

\title{The Homunculus Brain and Categorical Logic}
\author{Steve Awodey \\
Departments of Philosophy and Mathematics \\
Carnegie Mellon University \\
5000 Forbes Avenue \\
Pittsburgh, PA 15213 
\and
Michael Heller \\
Jagiellonian University \\
Copernicus Center for Interdisciplinary Studies\\
ul. Szczepa\'nska 1/5, 31-011 Cracow, Poland}

\date{\today}
\date{\today}
\maketitle

\begin{abstract}
The interaction between syntax (formal language) and its semantics (meanings of language) is one which has been well studied in categorical logic. The results of this particular study are employed to understand how the brain is able to create meanings. To emphasize the toy character of the proposed model, we prefer to speak of the homunculus brain rather than the brain per se. The homunculus brain consists of neurons, each of which is modeled by a category, and axons between neurons, which are modeled by functors between the corresponding neuron-categories. Each neuron (category) has its own program enabling its working, i.e. a theory of this neuron. In analogy to what is known from categorical logic, we postulate the existence of a pair of adjoint functors, called Lang and Syn, from a category, now called BRAIN, of categories, to a category, now called MIND, of theories. Our homunculus is a kind of ``mathematical robot'', the neuronal architecture of which is not important. Its only aim is to provide us with the opportunity to study how such a simple brain-like structure could ``create meanings'' and perform abstraction operations out of its purely syntactic program. The pair of adjoint functors Lang and Syn model the mutual dependencies between the syntactical structure of a given theory of MIND and the internal logic of its semantics given by a category of BRAIN. In this way, a formal language (syntax) and its meanings (semantics) are interwoven with each other in a manner corresponding to the adjointness of the functors Lang and Syn.  Higher cognitive functions of abstraction and realization of concepts are also modelled by a corresponding pair of adjoint functors.  The categories BRAIN and MIND interact with each other with their entire structures and, at the same time, these very structures are shaped by this interaction.
\end{abstract}

\section{Introduction: On the Computer Screen}
We were preparing a paper for publication. A phase portrait was nicely displayed on the computer screen. The network of trajectories represented a class of solutions to the equation we were interested in. At some points, called critical points, certain trajectories crossed each other. These points were important for our analysis. Some of the diagrams we worked with appeared later as figures in our publication \cite{GRG}. The figures had to be explained, so we decided to attach appropriate labels to some of the critical points. We attached the label ``stable saddle'' to one of them. No problem. Then we proceeded to attach the label ``unstable saddle'' to another one. But the label jumped up. We tried to fix it up, but it jumped down. Then we started laughing. After all, it is an unstable point!

Let us try to understand the situation. We were investigating an equation that (virtually) contains in itself its space of solutions (irrespectively of whether we explicitly know them or not). Through the suitable computer program and some ``electronic circuits'', which are activated by the program, this space of solutions is mapped into the phase portrait displayed on the computer screen. The diagram we see on the screen is certainly something more than just a picture. It does not simply show stable and unstable critical points; it also does what the abstract equation orders its solutions to do (labels jump up and down at instabilities).

Let us go a step forward. In fact, the phase portrait on the screen is a substitute of the world. For suppose that our equation ``describes'' (or better -- models) a mechanical system (e.g., a pendulum or oscillator).\footnote{The equations we considered in our publication referred to a cosmological situation.}  Then the unstable critical points of our equation correspond to physical situations in which the considered mechanical system behaves in an unstable way. We thus have, on the one hand, an equation (or a set of equations) or, more broadly, a mathematical theory and, on the other hand, a domain (or an aspect) of the physical world of which the considered mathematical theory is a model. Between the mathematical model and the domain (or aspect) of the physical world there is a mysterious correspondence -- a correspondence in the root-meaning of this word: both sides co-respond to each other. It is an active correspondence, and the activity goes both ways: it looks as if the domain of the world informed the theory about its own internal structure, and the theory answered by prescribing what the domain should do. And the domain does it. The equations prescribe what the world should do, and the world executes this. The equations and the world  are coupled with each other and act in unison.

And the screen on my computer? It is a part of the world. The program we have constructed reads the structure of the equations and executes what the equations tell it. And because of the coupling between the equations and the world, the computer does, in miniature, what does the world on its own scale. This is the reason why computers are so effective in our reading of the structure of the world.

There is another domain in which a formal structure reveals its effective power and produces real effects. Such processes occur in the brain. The formal structure in question consists of electric signals propagating along nerve fibres between neurons across synapses, and the world of meanings should be regarded as a product of this activity. The interaction seems to go both ways: the ``language of neurons'' (what happens in the brain) produces the meanings related to this language (in the mind), and the meanings somehow influence the architecture of neurons. 

It seems that in both these cases (mathematical laws and their effects in the real world, and the brain -- mind interactions) we meet two instances of the same working of logic where syntax (a formal structure), by effectively interacting with its semantics, produces real effects. This kind of interaction, although kept strictly on the level of logic (i.e. with no reference to processes in the real world), is well known in the categorical logic. In the present paper, we attempt to employ these achievements of categorical logic to try to understand the brain--mind interaction. 

The traditional terminology of brain and mind (irrespective of current trends in cognitive sciences to get rid of their conceptual load) seems especially well adapted to the present context in which general ideas are more important than structural details. Moreover,  to avoid too hasty associations with the human brain and to emphasize the toy character of the proposed model, we prefer to speak of the homunculus brain rather than just of brain.

The action of our argument develops along the following lines. Section 2 is a reminder on formal language, its syntax and semantics. Sections 3 and 4 briefly review those parts of categorical logic that refer to these concepts. Every category, call it $C$, has its internal logic, and if this logic is sufficiently rich, the category provides semantics for a certain formal theory $T$. Moreover, there exists a pair of adjoint functors, called Lang and Syn, from a category, called CATEGORIES, of categories belonging to a certain class  (for instance, coherent categories) to a category, called THEORIES, of theories and \textit{vice versa}, which describe mutual dependencies between the syntactical structure of $T$ and the internal logic of its semantics given by $C$. This is described in section 3. In this way, syntax and semantics are interwoven with each other in a manner corresponding to the adjointness of the functors Lang and Syn. This is explored in section 4.  In section 5, we consider a deep categorical duality between the syntactic category of a theory and its individual models and suggest a functional interpretation in terms of abstraction and realization of concepts, in anticipation of the cognitive interpretation to be introduced next.
In section 6, the category CATEGORIES becomes the category BRAIN. It constitutes a simple model of a homunculus' brain. Objects of this category are categories (belonging to a certain class); every such category models a neuron. Morphisms of this category model signals propagating along nerve fibres between neurons. The category THEORIES becomes the category MIND. Its objects are ``theories of neurons''; more precisely, if $C \in $ BRAIN, then its ``theory'' is Lang$(C)$ in MIND. Morphisms of this category are functors between the corresponding syntactic theories; more precisely, if $T_1, T_2 \in $ MIND, then the morphism between them is $\mathrm{Syn}(T_1) \to \mathrm{Syn}(T_2)$. The pair of adjoint functors Lang and Syn model the interaction between the syntax of ``theories'' and their semantics, i.e. the network of neurons. The categories BRAIN and MIND are indeed somehow related to what their names refer to, at least as far as homunculus' brain and mind are concerned.

Following the seminal paper of W.S McCulloch and W. Pitts \cite{McCulloch}, published as early as in 1943, which proposed using classical logic to model neural processes in the brain, there have been so many papers developing and modifying (with various logical systems) this idea, that to quote even a sample of them would be immaterial (for a relatively recent state of art see a short review \cite{Koch}). A. Ehresmann \cite{Ehresmann} claims that it was R. Rosen \cite{Rosen} who was the first to employ category theory to model biological systems. A series of works followed (a non-representative sample: \cite{Gomez,Healy,Mizraji,Naotsugu}) proposing  the use of various parts of category theory to model different aspects of the brain activity. In particular, adjoint functors were suggested to model ``a range of universal-selectionist mechanisms'' \cite{Ellerman}. However, we have not been able to find something similar to modeling the interaction between brain's language and its meaning anywhere.

\section{Syntax and Semantics}
In linguistics, syntax and semantics are regarded as parts of semiotics, the study of signs. Syntax studies relations between signs, and semantics relations between signs and what the signs refer to.\footnote{Sometimes one also distinguishes pragmatics which studies relations between signs and their users.} Syntactic properties are attributed to linguistic expressions entirely with respect to their shape (or form). Semantics, on the other hand, endows them with meaning by referring signs to what they signify. Logic adapts these ideas to its own needs. Since it is a formal science, the signs it considers should be elements of a formal language, and they cannot refer to anything external. Halvorson put it, ``But a formal language is really not a language at all, since nobody reads or writes in a formal language. Indeed, one of the primary features of these so-called formal languages is that the symbols don't have any meaning'' \cite{Halvorson2016}. This is why the meaning should be ``artificially'' constructed for them. The idea of how this should be done can best be seen in Tarski's prototype of this procedure \cite{Tarski}. If a sentence $s$, the truth of which we want to define, belongs to a language $L$ then the definition of $s$ should be formulated in a metalanguage $M$ with respect to the language $L$. And the metalanguage $M$ should contain a copy of $s$ so that anything one can say with the help of $s$ in $L$, can also be said in $M$. The definition of ``True'' should be of the form

\begin{center}
For all $x$, True$(x)$ if and only if $\varphi (x)$
\end{center}
with the condition that ``True'' does not occur in $\varphi $. Here $x$ stands for the copy of the sentence $s$ in the metalanguage $L$, and $\varphi(x)$ describes, also in $M$, the state of affairs of which the sentence $s$ in $L$ reports (for more details see \cite{Stanford,Gila}). A metalinguistic copy of $s$ could also be expressed as ``s'' (taken in quotes). In Tarski's own example:
\begin{center}
``It snows'' is true iff it snows.
\end{center}
For pedagogical reasons, this example is taken from colloquial language, but strictly speaking Tarski's definition refers to formal languages. The formal language $L$ has its own syntax (since it is a formal language), but is lacking its semantic reference. As we have seen, such a reference had to be constructed for it with the help of the metalanguage $M$.

Now, the idea is to improve the situation by looking for such a conceptual context in which a semantics for a given theory would arise in a more natural (or even spontaneous) way.

\section{Categorical Semantics}
To do so we must first define precisely what we mean by language. Since the definition must be precise, let us choose as an example the language of mathematics based on standard first order logic (which is enough for most of the usual mathematics). Many other languages may be formalized in a similar way. In such a language we distinguish:
\begin{itemize}
\item constants: $0, 1, 2,\ldots , a, b, c,\ldots ,$ and variables: $x, y, z,\dots $, which can be combined by primitive operations to give
\item terms, for example: $x+y, \, x^3, \ldots $ which, in turn, can be combined, with the help of primitive relations, such as $=, <, \leq, \dots $, to produce
\item formulae, for example: $x+y = z, \, x \leq y,\ldots $ which, in turn can be combined, with the help of the usual logical connectives and quantifiers, into
\item more complicated formulae.
\end{itemize}

To make the language more flexible and more adapted for concrete applications, we diversify its expressions into various types (called also sorts). In mathematics, we might use different letters for natural and real numbers, or different symbols for vectors an scalars. We say that, in both cases, we are using a two-typed language.  There may be languages with as many types as is needed.

What we need is not so much a language, but rather a theory. In mathematical logic theory is almost the same as language; it is a formal language aimed at axiomatizing a certain class of sentences. The concept of theory, as it is functioning in modern physics can, in principle, be regarded as the special case of the logical concept of theory, although in scientific practice theories are rarely formulated with the full logical rigor.  

Let then $T$ be a theory expressed in a multi-type language.  Such a theory is defined as consisting of the following data:
\begin{enumerate}
\item
A set of types $\{X_1, X_2, \ldots \, X, Y, \ldots\; \}$.
\item
A set of variables $\{x, y, z, \ldots ,\, x_1, x_2, x_3, \ldots \}$ with a type assigned to each variable.
\item
A set of function symbols with a type assigned to each domain and codomain of every function symbol; for instance, to the term $x+y$, with the variable $x_1$ of type $X_1$ and the variable $x_2$ of type $X_2$, there corresponds the function symbol $f: X_1 \times X_2 \to Y$, and the term $f(x_1,y_1) = x_1+x_2$ is of type $Y$.
\item
A set of relation symbols with a type assigned to each argument of every relation symbol; for instance, to the formula $x+y=z$, with the variable $x$ of type $X_1$, the variable $y$ of type $X_2$ and the variable $z$ of type $X_3$, there corresponds the relation symbol $R \subseteq: X_1 \times X_2 \times X_3$, and $R(x,y,z)$ is an atomic formula.
\item
A set of logical symbols.
\item
A set of axioms for a given theory build up from terms and relation symbols with the help of logical connectives and quantifiers, respecting types of all terms.
\end{enumerate}
This is, in fact, a purely syntactic definition of theory (for details  see \cite[pp. 344-348]{Borc}, \cite[pp. 527-530]{MacLaneMoerdijk}). Now, we want to create a semantics, i.e. a model, for a theory $T$. This is done by constructing a category $C_T$ which will serve us as such a model. The construction is almost obvious:
\begin{enumerate}
\item
each type of $T$ is an object of $C_T$, 
\item
for each function symbol $f$ with the types $A$ and $B$ of its domain and codomain in $T$, correspondingly, $f$ is a morphism from the object $A$ to the object $B$ in $C_T$,\footnote{Since $f$ is now in $C_T$ rather than in $T$, it should formally be denoted by a different symbol.}
\item
each variable is an identity morphism in $C_T$,
\item
for each relation symbol $R$ in $T$, its counterpart in $C_T$  is a subobject in $C_T$. Suppose $\phi $ is a subobject of an object $A$ in $C_T$ then, by analogy with the usual theory of sets, $\phi $ can be thought of as a collection of all things of type $A$ that verify $\phi $.
\end{enumerate}
This definition must be supplemented with all of the (first order) logic which is used to express axioms in $T$ (for details \cite{nLab2}). Roughly speaking, since formulae correspond to subobjects, and all subobjects of a given object are partially ordered by inclusions (they form a poset), the axioms can be expressed in terms of the order relation on the subobject poset in the category $C_T$.
The category, defined in this way, is appropriately called the categorical semantics for a theory $T$.

We have thus created (almost automatically!) a domain (the category $C_T$) the theory $T$ refers to. The internal architecture of the category $C_T$ exactly matches the logic involved in the theory $T$. 

Let us also mention that, \textit{vice versa}, having a (sufficiently rich) category $C'$, we can construct the formal theory $T'$ the logic of which matches the internal architecture of the category $C'$. This can be done by reading the above definition of the categorical semantics ``backward'', i.e. we regard objects of $C'$ as types of $T'$, identity morphisms of $C'$ as variables in $T'$, etc. The theory $T'$, reconstructed in this way from the category $C'$, is called internal logic of $C$. This entire process can be regarded as a functor, called Lang, from a category of categories, call it CATEGORIES, to a category of theories, call it THEORIES,
\begin{center}
Lang: CATEGORIES $\to $ THEORIES.
\end{center}
For the time being this definition remains informal since neither CATEGORIES nor THEORIES have been properly defined, but it will be done below. 

Let us start with a formal theory $T$. We now want to organize it into a category Syn$(T)$, called the syntactic category of $T$. It is done in the following way.

Let $\Gamma $ be a collection of type assertions, i.e. a collection of rules assigning a type to each term of a given theory, and $\Phi $ a collection of all well defined formulae of $T$. The pair $(\Gamma , \Phi )$ is called a context. It is a formalization of what in ordinary language one means by this term.

If $T$ is a type theory, its syntactic category, Syn(T), is defined as follows. Its objects are contexts $(\Gamma , \Phi )$ and its morphisms $(\Gamma , \Phi ) \to (\Delta , \Psi )$ are interpretations (or substitutions) of variables. The latter means that for each type, prescribed by $\Delta $, we must construct an expression of this type out of data contained in $\Gamma $. In general, this is done by substituting terms from $\Gamma $ for variables in $\Delta $. We must also present, for each assumption required by $\Delta $ (if there are any), a proof of this assumption from the assumptions contained in $\Gamma $ (for details see \cite{Fu,nLabSyn}). 

The category Syn(T), constructed in this way, is also called a category of contexts (for details see \cite{Fu,nLabSubs}). 

Since from a theory $T$ we have constructed the category Syn($T$), we can have a  functor,
\begin{center}
Syn: THEORIES $\to$ CATEGORIES
\end{center}
provided we define the categories THEORIES and CATEGORIES. We do this in the next section.

\section{Syntax -- Semantics Interaction}
Let us start with objects for both of these categories. It is obvious that they will be categories and theories, respectively. To have  workable categories, one must restrict the class of theories as candidates of being objects in THEORIES (and analogously for CATEGORIES). The criterion one follows is the kind of logic that underlines a given theory. It could be what logicians call: finite product logic, regular logic, coherent logic, geometric logic, etc. (as it could be expected, the internal logic of the corresponding semantic category will be of the corresponding kind, i.e. finite product logic, regular logic, etc.) \cite{nLab2}. For our further analysis it is irrelevant which one will be chosen. However, for the sake of concreteness we may think about coherent logic. Roughly speaking, this is a fragment of the first order logic which uses only the connectives $\wedge $ and $\vee $, and the existential quantifier. Large parts of mathematics can be formalised with the help of this logic. To this logic there correspond coherent theories and coherent categories. They will constitute objects of THEORIES and CATEGORIES, respectively.
Morphisms for CATEGORIES are obviously functors between corresponding categories; for instance coherent functors for coherent categories \cite{nLabCoherent}.  Let now $T_1$ and $T_2$ be objects in THEORIES. Morphism between $T_1$ and $T_2$, $T_1, \to T_2$, is a functor between their corresponding syntactic theories $\mathrm{Syn}(T_1) \to \mathrm{Syn}(T_2)$. Roughly speaking, this means that it is possible to express (to interpret) $T_1$ in terms of $T_2$ (for details and discussion see \cite{Halvorson2017}).\footnote{Strictly speaking, CATEGORIES is a 2-category (since its objects are categories and morphisms are functors), and THEORIES is a 2-category, in this case, called also a doctrine \cite{nLabDoctrine}.}

As a side remark let us notice that by studying the category THEORIES, we could learn ``how individual theories sit within it, and how theories are related to each other'' \cite[p. 413]{Halvorson2017}. This is nicely consonant with a newer trend in the philosophy of science to investigate the so-called inter-theory relations \cite{Batterman,Rosaler}.

A truly remarkable fact is that the functors Lang and Syn constitute a pair of adjoint functors. Let us explain precisely what this means.

Let us consider any pair of objects: $C$ of CATEGORIES and $T$ of THEORIES. Adjoint functors serve to compare them. However, they cannot be compared directly since they live in different categories. Adjoint functors serve to move each of them to the correct category so as to enable the comparison. Let us follow this process step by step \cite[pp. 148-153]{Simmons}.

Let us first consider the object $T$ which lives in THEORIES. We want to compare it with the object $C$ which lives in CATEGORIES. We thus move $C$ to THEORIES with the help of the functor Lang to obtain the object Lang($C$). We now make the comparison with the help of a suitable morphism,
$$
f: \mathrm{Lang}(C) \to T
$$
in THEORIES. We do the same starting with $C$ in CATEGORIES and $T$ in THEORIES, and compare $C$ with Syn$(T)$, 
$$
g: C \to \mathrm{Syn}(T)
$$
in CATEGORIES. To complete the definition of adjunction we demand that morphisms $f$ and $g$ should constitute a pair of bijections which is natural both in $C$ and $T$ (see below).

The above definition can be put into a concise form
\begin{equation}
\label{hom1}
\mathrm{THEORIES}(\mathrm{Lang}(\CC ), T) \cong {\mathrm{CATEGORIES}}(\CC , \mathrm{Syn}(T)),
\end{equation}
expressing an isomorphism between the right and left hand sides of this formula that is natural in $C$ and $T$. The latter condition says that when \CC \ varies in CATEGORIES and $T$ varies in THEORIES, the isomorphism between morphisms $\mathrm{Lang}(\CC ) \to T$ in THEORIES and $\CC \to \mathrm{Syn}(T)$ in CATEGORIES vary in a way that is compatible with the composition of morphisms in CATEGORIES and THEORIES, correspondingly, and with the actions of Lang and Syn on both these categories (see \cite[pp. 50-51]{Leinster}).\footnote{For a full definition of adjoint functors see any textbook on category theory.}

We should notice that in the above definition, in fact, we not only compare objects of two different categories, but rather categories themselves (objects $C$ and $T$ are any pair of objects). Moreover, comparing two categories we are not so much interested in their objects, but rather in morphisms between objects. This is clear from the fact that at the end, we have identified those morphisms of two categories that are pairwise naturally isomorphic among themselves.

As we can see, categorical logic does not simply creates a semantics for a given language, but shows that dependencies between them go both ways: in a sense, syntax and semantics create each other. More precisely, they condition each other through the adjointness relation.

%
%
\section{Realization and Abstraction}

There is another aspect of categorical logic that we shall make use of, and it may be seen as a mathematical description of the processes of abstraction and realization of concepts. The category $\mathrm{Syn}(T)$ representing a theory $T$ may be regarded as presenting a general \emph{concept}, of which the theory $T$ is a particular syntactic description.  For example, there is a theory $T_{Group}$ consisting of 
a single basic type $X$, and function symbols $* : X\times X\to X$ and $(-)^{-1} : X \to X$ and a constant $u : X$, together with the usual equations for groups as its axioms:
\begin{align*}
x*(y*z) &= (x*y)*z\\
x*u &= x\\
u*x &= x\\
x*x^{-1} &= u\\
x^{-1}*x &=u
\end{align*}
The syntactic category $\mathrm{Syn}(T_{Group})$ then represents the general concept of a group.  This concept can also be represented by another theory $T_{Group}'$ with a different choice of basic equations, or even a different choice of operations,\footnote{For example there is an axiomatization of groups using a single ternary operation in place of the two operations $x*y$ and $x^{-1}$).} as long as the resulting categories $\mathrm{Syn}(T_{Group})$  and $\mathrm{Syn}(T_{Group}')$ are equivalent.

A general concept may have many individual \emph{instances}; an instance of the concept of a group is, of course, just a particular group: a set $G$ of elements, equipped with functions interpreting the operations of multiplication and inverse, and satisfying the group equations.  A logician would call such an instance a \emph{model of the theory of groups}, but we shall avoid this over-worked term and refer to it instead as a \emph{realization} of the theory of groups.  A realization of a theory $T$ in any category $\CC$ is essentially the same thing as a functor $\mathrm{Syn}(T) \to \CC$ that preserves the relevant structure of the theory -- in the case of groups, the finite products $X\times X$.  (This is in fact the defining universal property of the syntactic category $\mathrm{Syn}(T)$.)  The realizations in the category $SET$, consisting of all sets and functions, are thus exactly what we called the \emph{instances} of the general concept of a group, namely groups.  

The standard category $GROUP$ of all groups and their homomorphisms, as usually defined in abstract algebra, is then essentially the same as the category of all such instances, that is, the category $\mathrm{REAL}(\mathrm{Syn}(T_{Group}), SET)$ of all $SET$ realizations, i.e.\ (structure-preserving) functors, where the morphisms are just natural transformations of such functors (that these correspond exactly to group homomorphisms is not trivial).  In this way, for any general concept $\mathrm{Syn}(T)$ corresponding to a theory $T$ we can define the category of its $SET$ realizations, 
\[
\mathrm{REAL}(T) =_{df} \mathrm{REAL}(\mathrm{Syn}(T), SET),
\]
which may be viewed as the category of instances of the concept $\mathrm{Syn}(T)$.

Now an amazing and mathematically deep fact emerges, which can only be seen usng the tools of categorical logic: from the category  $\mathrm{REAL}(T) $ of all instances of the concept presented by $T$, one can actually recover the general concept $\mathrm{Syn}(T)$.  Indeed, for any structured category $\mathcal{R}$ of the same kind as $\mathrm{REAL}(T)$ (we will say a bit more about the condition ``of the same kind'' below), one can consider all of the \emph{continuous} functors $f : \mathcal{R} \to SET$; these may be regarded as ``images'' or ``abstractions'' of the (generalized) realizations in $\mathcal{R}$.  The category of all such abstractions $\mathrm{ABSTRACT}(\mathcal{R}, SET)$ (again, with natural transformations as morphisms) may be called the \emph{abstract} of $\mathcal{R}$, and written simply 
$$
\mathrm{ABSTRACT}(\mathcal{R}) =_{df}  \mathrm{ABSTRACT}(\mathcal{R}, SET).
$$
A similar construction that the reader may know is the ring $\mathcal{C}(X) = \mathcal{C}(X, \mathbb{R})$ of continuous, real-valued functions on a space $X$. The noteworthy fact that we mentioned above is this: if for $\mathcal{R}$ we take a category $\mathrm{REAL}(T) $ of realizations of a theory $T$, then the abstract of $\mathrm{REAL}(T) $, consisting of all ``abstractions'' $\mathrm{REAL}(T) \to SET$, will be the associated concept $\mathrm{Syn}(T)$.\footnote{Under suitable assumptions, and up to the relevant notion of equivalence, of course; see \cite{Awodey2019} for the general theory.}   Thus \emph{the abstraction of the realizations of a concept is the concept itself}.  We can even summarize this briefly by saying that \emph{All concepts are abstract}, since every concept is the abstraction of its realizations.  More generally, for any suitable category $\mathcal{R}$, the category $\mathrm{ABSTRACT}(\mathcal{R})$ of all continuous functors $f : \mathcal{R} \to SET$ (the ``abstractions'' of $\mathcal{R}$) is a general concept, of which $\mathcal{R}$ is either the category of realizations, or an approximation thereof.  

The general correspondence is given by a (contravariant!) adjunction between the \emph{functors} of Realization and Abstraction which relate these operations; schematically,
\[
\xymatrix{
\mathrm{CONCEPTS}  \ar@/^4ex/ [r]^{\mathrm{Realization}} & \ar@/^4ex/ [l]^{\mathrm{Abstraction}} \mathrm{INSTANCES}^{op}
}
\]

Here $\mathrm{CONCEPTS}$ is the category consisting of all ``conceptual'' categories $\mathrm{Syn}(T)$ and their (relative) ``realizations'', i.e.\ functors  $\mathrm{Syn}(T) \to \mathrm{Syn}(T')$, and the functor of Realization is defined by taking realizations in $SET$,
\begin{align*}
\mathrm{Realization}(\mathrm{Syn}(T)) = \mathrm{REAL}(\mathrm{Syn}(T), SET)\,.
\end{align*}
which we also called the category of ``instances'' of the concept. 

And $\mathrm{INSTANCES}$ is the category consisting of all (generalizaed) categories of instances $\mathcal{R}$ (such as the categories $GROUP$, $RING$, etc.) with their ``continuous'' functors  $\mathcal{R} \to \mathcal{R'}$, and the functor of Abstraction is defined by taking continuous functors into $SET$, 
\begin{align*}
\mathrm{Abstraction}(\mathcal{R}) = \mathrm{ABSTRACT}(\mathcal{R}, SET)\,.
\end{align*}
which we called ``abstractions'' of the category $\mathcal{R}$. 

Let us consider a simple example!  Propositional logic consists of basic propositional variables $x, y, z, \dots$ and constants $\top, \bot$, which can be made into formulae using the usual propositional connectives $\neg z\,, x \wedge y\,, x\vee y\,, x\Rightarrow z$, and which are assumed to satisfy the usual logical laws, such as $x \wedge (y\vee z) = (x \wedge y)\vee (x \wedge z)$\,, $\neg\neg x = x$, etc.  A \emph{theory} $T$ in this simplified case is just a set of propositional letters $V = \{p_1, p_2, \dots, p_n\}$ (regarded a 0-ary relation symbols), and a list of propositional formulae $A = \{\alpha_1, \alpha_2, \dots, \alpha_m\}$ built up from these letters, as the axioms of the theory.  There are no types, typed variables, or function symbols (or rather, there is a single, implicit type $1$), and the logical symbols are just the propositional connectives.

The \emph{syntactic category} $\mathrm{Syn}(T)$, representing the ``concept'',  is then the Boolean algebra $F(V)/A$ obtained as the free Boolean algebra $F(V)$ on the variables $V$ as generators, quotiented by the filter generated by the axioms $A$.  This ``concept'' associated to the propositional theory $T = (V,A)$ is independent of the particular syntactic presentation $(V,A)$.  A \emph{realization} of $T$ is then a boolean homomorphism $F(V)/A \to 2$, where $2 = \{0,1\}$ is the Boolean algebra of truth values.  Thus such a realization is just a truth-value assignment to the variables in $V$, in such a way that the ``conditions'' in $A$ are all satisfied, i.e.\ the elements $a\in A$ are all taken to the value ``true'' (in other words, a ``model'' of the propositional theory $T$).  For instance, if the theory $T$ is $V= \{x, y\}$ and $A = \{x\vee y, \neg(x\wedge y)\}$, then a realization would be an assignment of $x$ to an actual sentence $p$, and $y$ to one $q$, such that only one of $p$ and $q$ is true (or more formally, a direct assignement of such truth values, by-passing the actual sentences).  Under our description above, such a realization is an instance of the general concept $F(V)/A$.

Now, such realizations are exactly the points of the \emph{Stone space} $\mathrm{Stone}(F(V)/A)$, the topological space associated to the Boolean algebra $F(V)/A$ under the celebrated Stone duality theorem \cite{Johnstone} -- which is in fact the ``propositional logic'' case of the categorical logical duality that we are considering here.  Formally, the points of $\mathrm{Stone}(F(V)/A)$ are prime filters in $F(V)/A$, and the topology has basic open sets determined by the elements of $F(V)/A$.  The Boolean algebra $F(V)/A$ can be recovered from this space $\mathrm{Stone}(F(V)/A)$ as the algebra of continuous functions $\mathrm{Stone}(F(V)/A)\to \mathsf{2}$ into the discrete space $\mathsf{2}$, with the pointwise Boolean operations.  These \emph{abstractions} of $\mathrm{Stone}(F(V)/A)$ form a Boolean algebra $\mathrm{Bool}(\mathrm{Stone}(F(V)/A))$ which, by Stone duality, is isomorphic to $F(V)/A$,
\[
\mathrm{Bool}(\mathrm{Stone}(F(V)/A))\ \cong\ F(V)/A\,.
\]  
Indeed, for any (not necessarily Stone) space $X$, we can form the Boolean algebra $\mathrm{Bool}(X)$ of continuous functions $X \to \mathsf{2}$, and the original space $X$ will then map canonically to $\mathrm{Stone}(\mathrm{Bool}(X))$, giving the ``best approximation'' of $X$ by a Stone space.

In the general case, in categorical logic we consider many other fragments of logic --- propositional, equational, coherent, first-order --- and for each such subsystem there is an associated Realization-Abstraction adjunction between theories, and the concepts they represent, on the one hand, and their realizations by instances of these concepts, on the other.  The propositional theories just considered give rise to Stone duality \cite{Johnstone}; equational theories (like groups) give rise to Lawvere duality \cite{ALR}; coherent and first-order logic are treated by analogous duality theories developed by Makkai and others \cite{Makkai,AF}.  In each diffferent case, the associated notion of structured category, structure-preserving functor, continuous functor, etc., is suitably adapted to the respective situation.  Many of these logical dualities are discussed from the standpoint of categorical logic in the paper \cite{Awodey2019}.

\section{Categories BRAIN and MIND}
So far everything that has been said has merely been a reminder of standard and well known things. From now on, everything will be hypothetical and highly simplified. The bold and maximally simplified hypothesis is that neurons in the  brain can be modeled as categories, the internal logic of which is sufficiently complex (yet manageable). Of course, our inspiring motive is the human brain and in constructing our model we shall try to imitate what is going on it; however, being conscious of our simplified and highly idealized assumptions, we prefer to speak about a homunculus brain. Our homunculus is a kind of ``mathematical robot'', the aim of which is to provide us with the opportunity to study how such a simple brain-like structure could ``create meanings'' out of its purely syntactic program. Our other drastically simplifying assumption consists in systematically ignoring all of the brain's functions and processes that are not directly related to the proposed syntax--semantics relationship.

As it is well known, neurons communicate through signals transmitted via: presynaptic (source) neuron -- axon --  synapse -- dendrite -- postsynaptic (target) neuron, and this \textit{via} is unidirectional. In our homunculus model, these transmission processes will be regarded as functors between categories (neurons). 

Let us consider the category CATEGORIES, which we now aptly call BRAIN. Its objects are categories modeling neurons, and morphisms are functors between these categories.

We thus assume that each neuron in the homunculus brain is represented by a category (belonging to a certain class of categories; in the following we shall simply say that a neuron is a category). At the moment, we are not interested which biological mechanisms implement this assumption. Everything that counts in this model is the assumption that neurons consist of collections of objects and morphisms satisfying conditions from the category definition. We should have in mind that these simple conditions might lead to highly complicated structures.

Morphisms (arrows) in the category CATEGORIES are functors between object-categories, that is to say axons through which neurons communicate with each other. The crucial thing is that they must satisfy the usual conditions for morphisms: composition of morphisms, its associativity, the existence of identity morphisms. With the latter there is no problem: no output from a neuron counts as its identity morphism. To check whether two other conditions are verified in the human brain would require going deeper into the neural structure of our brain. In the case of the homunculus brain, this is not necessary. Since the homunculus is of our construction, we simply assume that synapses in its brain well compose and do so in the associative way.

The next step seems obvious. Each neuron (modeled as a category $C \in $ BRAIN) has its own program enabling its working, i.e. an internal logic underlying this program. We thus can define a counterpart of Lang$(C)$ which is a ``theory'' of this neuron. It is reasonable to claim that it is an object of the category THEORIES which we now call MIND, and the functor Lang: BRAIN $\to $ MIND is defined in analogy to that between CATEGORIES and THEORIES.

What about the morphisms between such objects? We proceed in strict analogy with what has been done in THEORIES. Let now $T_1$ and $T_2$ be objects in MIND, a morphism between them, $T_1, \to T_2$, is a functor between their corresponding syntactic theories, i.e. $\mathrm{Syn}(T_1) \to \mathrm{Syn}(T_2)$, where the functor Syn: MIND $\to $ BRAIN is defined in analogy to that between THEORIES and CATEGORIES.

The analogy is only apparently straightforward. In fact, it is based on a huge extrapolation, and as such highly hypothetical, but it is worth exploring it since the problem at stake deserves even a higher risk. By pursuing this analogy we could claim that also in this case the functors Lang and Syn are adjoint functors. If so, we have a very interesting conjunction between brain and mind; it is interesting even if brain and mind are modeled by such a naive construction. 

Neurons, their interactions and programs underlying their working  are, in contrast with abstract categories like CATEGORIES and THEORIES, real things, at least in the homunculus world, and we are entitled to suppose that the functors Lang and Syn between Brain and Mind really do what they formally signify (like our phase portrait on the computer screen really did what the program told it to do). 

Roughly speaking the functor Lang provides a collection of theories (mind) for a collection of neurons (brain), and the functor Syn transfers the syntax of these theories to the network of neurons. The action of these two functors is adjoint; consequently it determines a strict interaction between BRAIN and MIND. Let $C$ be any object (a neuron) in BRAIN and $T$ any object (the theory of this neuron) in MIND, then equation (\ref{hom1}) assumes the form
\begin{equation}
\label{hom2}
\mathrm{MIND}(\mathrm{Lang}(\CC ), T) \cong {\mathrm{BRAIN}}(\CC , \mathrm{Syn}(T)).
 \end{equation}
The natural isomorphism $\cong $ appearing in this equation is crucial. It states that when we go from neuron to neuron as objects in BRAIN, and their corresponding theories vary in THEORIES, then the isomorphism between morphisms $\mathrm{Lang}(\CC ) \to T$ in MIND and $\CC \to \mathrm{Syn}(T)$ in BRAIN varies in a way that is compatible with the composition of morphisms in BRAIN and MIND, correspondingly, and with the actions of the functors Lang and Syn (see \cite{Leinster}).\footnote{For a full discussion of the role of the naturality condition in the definition of adjoint functors see any textbook on category theory.} 
%
Finally, the ``higher'' cognitive functions of abstraction and realization of concepts are modelled by a corresponding adjunction between the assocated functors $\mathrm{Abstraction}$ and $\mathrm{Realization}$ relating these categories BRAIN and MIND. 
%
We could summarise the situation by saying that the categories BRAIN and MIND interact with each other with their entire structures and, at the same time, these very structures are shaped by this interaction.

\section{A Comment}
The interactions between syntax and semantics are omnipresent both in our everyday conversations and in various forms of practicing science. The world around us is full of meanings and our attempts to decipher them. Science could be regarded as a machine to produce signs, through experimentation and critical reasoning, and  extracting from combinations of them information about the structure of the world. Logicians put a lot of effort to make the syntax -- semantics interaction precise. As we have seen in section 2, despite the fact that formal languages are lacking any external references, it was possible to create semantical references for them  by cleverly exploiting the relation between language and its metalanguage. In categorical logic the situation has improved. Any formal theory $T$, generates via the functor Syn the category $\mathrm{Syn}(C)=C_T$ of which it is a theory, i.e. $C_T$ provides a ``natural'' semantics for $T$. And \textit{vice versa}, any (sufficiently rich) category $C'$, via the functor Lang, generates its own theory Lang$(C')=T'_{C'}$ which constitutes the internal logic of $C'$. It is interesting to notice that  $T_{C_T}$ does not coincide with $T$, they are only Morita equivalent. Here, we shall not go into technical details; it is enough to say that two Morita equivalent theories could be regarded as two interpretations of the same theory \cite{Halvorson2016}.

The fact that $T_{C_T}$ does not coincide with $T$ is a consequence of the fact that the functors Lan and Syn are not mutually inverse functors but constitute a pair of adjoint functors. This in turn implies that in categorical logic the interaction between syntax and semantics is skillfully complex, with creative influences coming both ways.

All the above discussed properties of the syntax -- semantics interaction are obviously preserved if applied to the categories BRAIN and MIND. There is only one big difference: now ``neurons and their theories'' are real things (although in a highly idealised version of the homunculus world). Nevertheless, the situation is not so different from the one which we can observe in many empirical sciences, in which some abstract mathematical structures model some real processes (always more or less idealised). We should not be surprised that the method of mathematical modeling  works when applied to our cognitive processes, but rather that mathematical structures not only describe the real world (whether it is our brain or the world of physics), but that they are also effectively acting in it (like in the little arrow on the computer screen).


\begin{thebibliography}{cc}
\bibitem{ALR}
Adamek, J.,  Lawvere, F.W.,  Rosicky, J., On the duality between varieties and algebraic theories, 
\textit{Algebra Universalis} 49, 2003, pp.~35--49.
\bibitem{Awodey2019}
Awodey, S., Sheaf representations and duality in logic, 2019. \url{https://www.andrew.cmu.edu/user/awodey/preprints/lambek.pdf}.
\bibitem{AF}
Awodey, S., Forssell, H., First-Order Logical Duality, \textit{Annals of Pure and Applied Logic} 164, 2012. arXiv:1008.3145 [math.LO]
\bibitem{Barrett2015}
Barrett, T., Halvorson, H., Morita Equivalence, arXiv:1506.04675 [math.LO]
\bibitem{Batterman}
Batterman, R. Intertheory Relations in Physics, The Stanford Encyclopedia of Philosophy, \url{https://plato.stanford.edu/archives/fall2016/entries/physics-interrelate}
\bibitem{Borc}
Borceux, F., \textit{Handbook of Categorical Algebra, vol. 3: Categories of Sheaves}, Cambridge University Press, Cambridge, 1994.
\bibitem{nLabCoherent}
Coherent Functors, nLab, 2011, \url{https://ncatlab.org/nlab/show/coherent+functor}
\bibitem{nLabDoctrine}
Doctrine, in nLab, \url{https://ncatlab.org/nlab/show/doctrine}
\bibitem{Ehresmann}
Ehresmann, A., Applications of Categories to Biology and Cognition, in: \textit{Categories for the Working Philosopher}, ed. by E. Landry, Oxford University Press, Oxford, 2017, pp.358-380. 
\bibitem{Ellerman}
Ellerman, D., On Adjoint and Brain Functors, arXiv:1508.04036v1 [math.CT]
\bibitem{Fu}
Fu, Yuchen, Category Theory, Topos and Logic: A Quick Glance, \url{http://charlesfu.me/repository/topos.pdf}
\bibitem{Gomez}
Gomez, J., Sanz, R., \textit{Modeling Cognitive Systems with
Category Theory}, Universidad Politecnica de Madrid, 2009.
\bibitem{Halvorson2016}
Halvorson, H., Scientific Theories, in: \textit{Oxford Handbook of the Philosophy of Science}, ed. by P. Humphreys, Oxford University Press, Oxford, 2016, pp. 585-608.
\bibitem{Halvorson2017}
Halvorson, H., Tsementzis, D., Categories of Scientific Theories,  in: \textit{Categories for the Working Philosopher}, ed. by E. Landry, Oxford University Press, Oxford, 2017, pp. 402-429.
\bibitem{Healy}
Healy, M.J., Caudell, T.P., Ontologies and Worlds in Category Theory: Implications for Neural Systems, \textit{Axiomathes} 16, 2006, 165-214.
\bibitem{Stanford}
Hodges, W., Tarski's Truth Definitions, \textit{The Stanford Encyclopedia of Philosophy} (Fall 2018 Edition), \url{https://plato.stanford.edu/archives/fall2018/entries/tarski-truth/}.
\bibitem{nLab2}
Internal Logic, nLab, 2017, \url{https://ncatlab.org/nlab/show/internal+logic}
\bibitem{Johnstone}
Johnstone, P.T.J, \textit{Stone Spaces}, Cambridge Studies in Advanced Mathematics 3, Cambridge University Press, 1982.
\bibitem{Koch}
Koch, Ch., Computation and the Single Neuron, \textit{Nature} 385, 1997, 207-210.
\bibitem{Leinster}
Leinster, T., \textit{Basic Category Theory}, Cambridge University Press, Cambridge, 2014.
\bibitem{MacLaneMoerdijk}
Mac Lane, S., Moerdijk, I., \textit{Sheaves in Geometry and Logic}, Springer, New York -- Berlin -- Heidelberg, 1992.
\bibitem{Makkai}
Makkai, M., Stone duality for first order logic, \textit{Adv. Math.} 65, 1987, pp.~97--170.
\bibitem{McCulloch}
McCulloch, W.S. and Pittts, W., A Logical Calculus of the Ideas Immanent in Nervous Activity, \textit{Bulletin of Mathematical Biophysics} 5, 1943, 115-133.
\bibitem{Mizraji}
Mizraji, E., Lin, J., Logic in a Dynamic Brain, \textit{Bulletin of Mathematical Biology} 73, 2011, 373-97, doi: 10.1007/s11538-010-9561-0
\bibitem{Naotsugu}
Naotsugu Tsuchiy, Shigeru Taguchi, Hayato Saigo, Perspective Using Category Theory to Assess the Relationship between Consciousness and Integrated Information Theory, \textit{Neuroscience Research} 107, 2016, 1–7
\bibitem{Rogers}
Rogers, R., \textit{Mathematical Logic and Formalised Theories}, North Holland, Amsterdam--London, 1971.
\bibitem{Rosaler}
Rosaler, J., Inter-Theory Relations in Physics: Case Studies from Quantum Mechanics and Quantum Field Theory, arXiv:1802.09350 [quant-ph]
\bibitem{Rosen}
Rosen, R., Organisms as Causal Systems which are not Mechanisms: an Essay into  the Nature of Complexity, \textit{Theoretical Biology and Complexiity} 1985, 165-203.
\bibitem{Gila}
Sher, G., What is Tarski's Theory of Truth? \textit{Topoi} 18, 1999, 149–166.
\bibitem{Simmons}
Simmons, H., \textit{An Introduction to Category Theory}, Cambridge University Press, Cambridge, 2011.
\bibitem{nLabSubs}
Substitution, in nLab, \url{https://ncatlab.org/nlab/show/substitution}
\bibitem{nLabSyn}
Syntactic Category, in nLab, \url{
https://ncatlab.org/nlab/show/syntactic+category}
\bibitem{Tarski}
Tarski, A., \textit{Poj\c{e}cie prawdy w j\c{e}zykach nauk dedukcyjnych}, Towarzystwo Naukowe Warszawskie, Warszawa, 1933 (in Polish).
\bibitem{nLabTh}
Theory, nLab, \url{https://ncatlab.org/nlab/show/theory}
\bibitem{Tsem}
Tsementzis, D., A Synthetic Characterization of Morita Equivalence, arXiv: 1507.02302
\bibitem{GRG}
Woszczyna, A., Heller M., Is a Horizon-free Cosmology Possible?, \textit{General Relativity and Gravitation} 22, no 12, 1990, 1367-1386.

\end{thebibliography}
\end{document}